%% file: 00-main.tex
\documentclass[runningheads]{llncs}
\usepackage{graphicx}
\usepackage{comment}
\usepackage{amsmath,amssymb} 
\usepackage{color}
\usepackage{caption,subcaption}

\newcommand{\taohu}[1]{\textcolor{blue}{[\textbf{Vincent}: #1]}}
\newcommand{\pengwan}[1]{\textcolor{green}{[\textbf{PW}: #1]}}
\newcommand{\psmm}[1]{\textcolor{orange}{[\textbf{PM}: #1]}}

\usepackage{bibunits}
\usepackage{amsmath}

\usepackage{sidecap}

\newcommand{\etal}{\textit{et al}.}
\newcommand{\ie}{\textit{i}.\textit{e}.}
\newcommand{\eg}{\textit{e}.\textit{g}.}

\usepackage{comment}
\usepackage{multirow}
\usepackage{algorithm,algpseudocode}
\usepackage{url}
\usepackage{bbding}
\usepackage{verbatim}
\usepackage{wasysym} 
\usepackage{xspace}
\usepackage{tabularx}

\usepackage{booktabs}  
\usepackage[table,x11names]{xcolor}
\definecolor{mygray}{gray}{0.95}
\definecolor{myred}{rgb}{1.0, 0.0, 0.0}

\usepackage{caption}

\usepackage{cite} 

\makeatletter
\newcommand{\printfnsymbol}[1]{%
  \textsuperscript{\@fnsymbol{#1}}%
}
\makeatother

\begin{document}
\pagestyle{headings}
\mainmatter
\def\ECCVSubNumber{300}  

%
%
%
%
\title{Localizing the Common Action \\Among a Few Videos}
%

\titlerunning{Localizing the Common Action Among a Few Videos}
%
%
\author{Pengwan Yang\inst{1,2}\thanks{Equal contribution. \email{\{yangpengwan2016, taohu620\}@gmail.com}} \and Vincent Tao Hu\inst{2}\printfnsymbol{1} \and Pascal Mettes\inst{2} \and Cees G. M. Snoek\inst{2}}
\authorrunning{P. Yang et al.}
%
\institute{Peking University, China \and
University of Amsterdam, the Netherlands\\
}

\maketitle

\begin{abstract}
This paper strives to localize the temporal extent of an action in a long untrimmed video. Where existing work leverages many examples with their start, their ending, and/or the class of the action during training time, we propose few-shot common action localization. The start and end of an action in a long untrimmed video is determined based on just a hand-full of trimmed video examples containing the same action, without knowing their common class label. To address this task, we introduce a new 3D convolutional network architecture able to align representations from the support videos with the relevant query video segments. The network contains: (\textit{i}) a mutual enhancement module to simultaneously complement the representation of the few trimmed support videos and the untrimmed query video; (\textit{ii}) a progressive alignment module that iteratively fuses the support videos into the query branch; and (\textit{iii}) a pairwise matching module to weigh the importance of different support videos. Evaluation of few-shot common action localization in untrimmed videos containing a single or multiple action instances demonstrates the effectiveness and general applicability of our proposal.
\\
\textbf{Code:}  \url{https://github.com/PengWan-Yang/commonLocalization} 

\end{abstract}

%
%


\begin{bibunit}
\input{0_intro}

\input{1_related}

\input{2_method}
\input{3_experiment}

\input{4_conclusion}
{\small
\bibliographystyle{ieee_fullname}
\bibliography{5_egbib}
}
\end{bibunit}

\newpage

\begin{bibunit}
\appendix
\addcontentsline{toc}{section}{Appendices}
\section*{APPENDICES}
\input{6_supp}

\end{bibunit}

\end{document}

%% file: 0_intro.tex
\section{Introduction}
The goal of this paper is to localize the temporal extent of an action in a long untrimmed video. This challenging problem~\cite{Duchenne2009, oneata2013fisher} has witnessed considerable progress thanks to deep learning solutions, \eg \cite{shou2016temporal, gao2017iccv, lin2018bsn}, fueled by the availability of large-scale video datasets containing the start, the end, and the class of the action~\cite{idrees2017thumos,caba2015activitynet,damen2018epic}. Recently, weakly-supervised alternatives have appeared, \eg~\cite{ wang2017untrimmed, paul2018wtalc, nguyen2019iccv, yang2017common, bojanowski2014weakly,kumar2017hide, kuehne2019hybrid,JainCVPR20}. They avoid the need for hard to obtain start and end time annotations, but still require hundreds of videos labeled with their action class. In this paper, we also aim for a weakly-supervised setup, but we avoid the need for any action class labels. We propose few-shot \textit{common} action localization, which determines the start and end of an action in a long untrimmed video based on just a hand-full of trimmed videos containing the same action, without knowing their common class label. 

\begin{figure*}[t]
	\centering
	\includegraphics[width=0.946\linewidth]{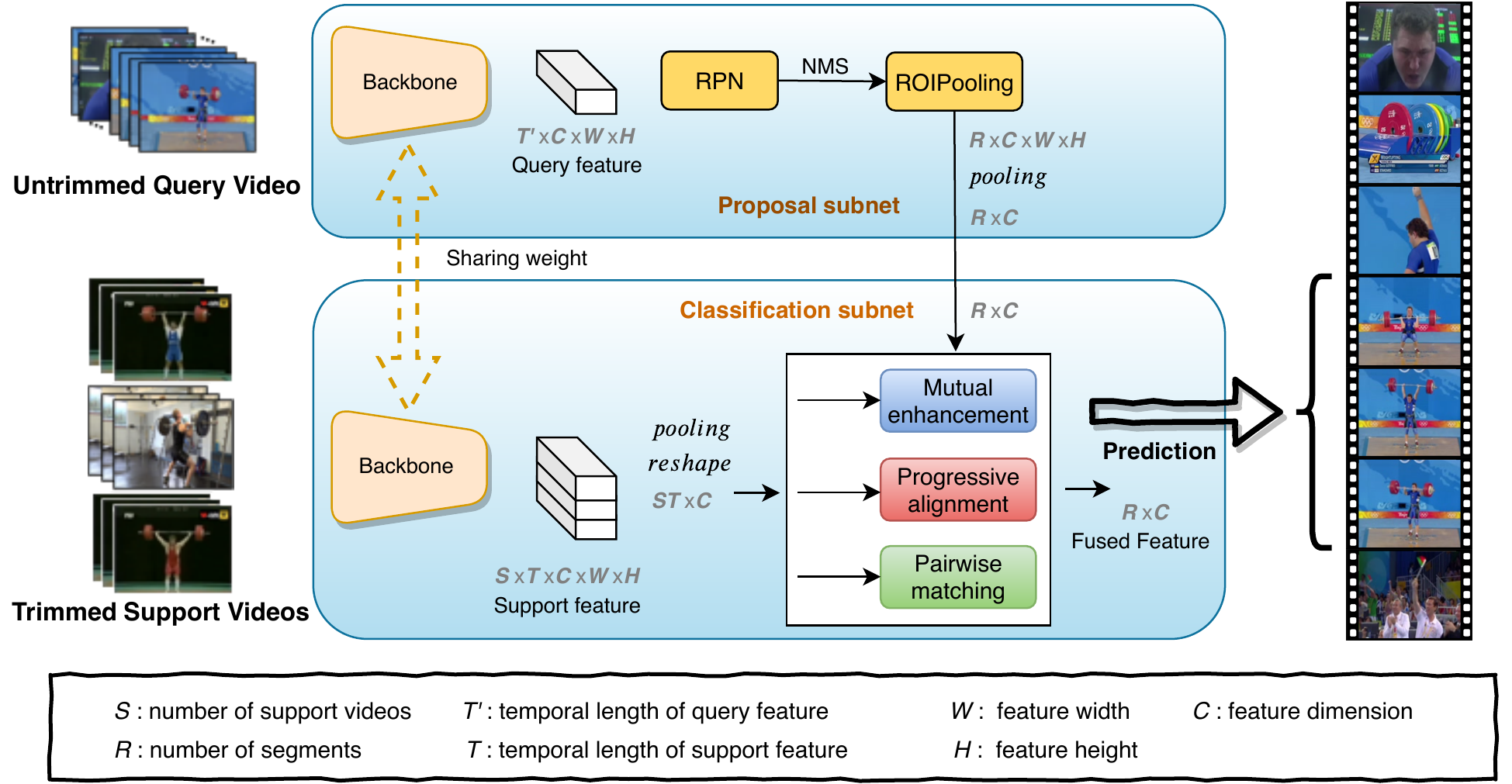}
	\caption{\textbf{Common action localization} in an untrimmed query video from three trimmed support videos during inference. 
	%
	%
	The action is localized in the query video based on the common action in the support videos.} 
	\label{fig:architecture}
\end{figure*}

We are inspired by recent works on few-shot object detection~\cite{dong2018few,sawatzky2018ex,silco2019,shaban2019learning}. Dong~\etal~\cite{dong2018few} start from a few labeled boxes per object and a large pool of unlabeled images. Pseudo-labels for the unlabeled images are utilized to iteratively refine the object detection result. Both Shaban~\etal~\cite{shaban2019learning} and Hu \etal~\cite{silco2019} further relax the labeling constraint by only requiring a few examples to contain a common object, without the strict need to know their class name. Hu \etal~\cite{silco2019} introduce two modules to reweigh the influence of each example and to leverage spatial similarity between support and query images. We also require that our few examples contain a common class and we adopt a reweighting module. Different from Hu~\etal, we have no module to focus on masking objects spatially in images. Instead, we introduce three alternative modules optimized for localizing actions temporally in long untrimmed videos, as illustrated in Figure~\ref{fig:architecture}. 

We make three contributions in this work. First, we consider common action localization from the few-shot perspective. All we require is that the few trimmed video examples share a common action, which may be obtained from social tags, hash tags or off-the-shelve action classifiers. Second, we propose a network architecture for few-shot common action localization, along with three modules able to align representations from the support videos with the relevant query video segments. The mutual enhancement module strengthens the representations of the query and support representations simultaneously by building upon non-local blocks~\cite{nl}. The progressive alignment module iteratively integrates the support branch into the query branch. Lastly, the pairwise matching module learns to weigh the importance of different support videos. As a third contribution, we reorganize the videos in ActivityNet1.3~\cite{caba2015activitynet} and Thumos14~\cite{idrees2017thumos} to allow for experimental evaluation of few-shot common action localization in long untrimmed videos containing a single or multiple action instances.

%% file: 1_related.tex
\section{Related work}

\textbf{Action localization from many examples.}
Standard action localization is concerned with finding the start and end times of actions in videos from many training videos with labeled temporal boundaries~\cite{buch2017sst,shou2016temporal,escorcia2016daps}.
A common approach is to employ sliding windows to generate segments and subsequently classify them with action classifiers~\cite{shou2016temporal,gao2018ctap,wang2014action,dai2017temporal,yang2018one}.
Due to the computational cost of sliding windows, several approaches model the temporal evolution of actions and predict an action label at each time step~\cite{escorcia2016daps,ma2016learning,singh2016multi,yeung2016end}. The R-C3D action localization pipeline~\cite{xu2017r} encodes the frames with fully-convolutional 3D filters, generates action proposals, then classifies and refines them. In this paper, we adopt the proposal subnet of R-C3D to obtain class-agnostic action proposals. In weakly-supervised localization, the models are learned from training videos without temporal annotations. They only rely on the global action class labels~\cite{nguyen2018weakly,wang2017untrimmed,nguyen2019iccv}. Different from both standard and weakly-supervised action localization, our \textit{common} action localization focuses on finding the common action in a long untrimmed query video given a few (or just one) trimmed support videos without knowing the common action class label, making our task class-agnostic. Furthermore, the videos used to train our approach contain actions that are not seen during testing.

\textbf{Action localization from few examples.}
Yang \etal~\cite{yang2018one} pioneered few-shot \textit{labeled} action localization, where a few (or at least one) positive labeled and several negative labeled videos steer the localization via an end-to-end meta-learning strategy. 
It relies on sliding windows to swipe over the untrimmed query video to generate fixed boundary proposals. 
Rather then relying on a few positive and many negative action class labels, our approach does not require any predefined positive nor negative action labels, all we require is that the few support videos have the same action in \emph{common}. Moreover, we propose a network architecture with three modules that predicts proposals of arbitrary length from commonality only.

\textbf{Action localization from one example.}
Closest to our work is video re-localization by Feng \etal~\cite{feng2018video}, which introduces localization in an untrimmed query video from a \textit{single} unlabeled support video. They propose a bilinear matching module with gating functions for the localization. Compared to video relocalization, we consider a more general and realistic setting, where more than one support video can be used. Furthermore, we consider untrimmed videos of longer temporal extent and we consider action localization from a single frame.
To enable action localization under these challenging settings, we introduce modules that learn to enhance and align the query video with one or more support videos, while furthermore learning to weigh individual support videos. We find that our proposed common action localization formulation obtains better results, both in existing and in our new settings.

\textbf{Action localization from no examples.}
Localization has also been investigated from a zero-shot perspective by linking actions to relevant objects~\cite{jain2015objects2action,kalogeiton2017joint,mettes2017spatial}. Soomro~\etal~\cite{soomro2017unsupervised} tackle action localization in an unsupervised setting, where no annotations are provided overall. While zero-shot and unsupervised action localization show promise, current approaches are not competitive with (weakly-)supervised alternatives, hence we focus on the few-shot setting.

%% file: 2_method.tex
\section{Method}

\subsection{Problem description}
For the task of few-shot common action localization, we are given a set of trimmed support videos $S_{c}^N$, where $N$ is small, and an untrimmed query video $Q_{c}$. Both the support and query videos contain activity class $c$, although its label is not provided. The goal is to learn a function $f(S_{c}^N, Q_{c})$ that outputs the temporal segments for activity class $c$ in the query video. The function $f(\cdot, \cdot)$ is parametrized by a deep network consisting of a support and query branch. During training, we have access to a set of support-query tuples $T = \{(S_l^N,Q_l)\}_{l \in \mathcal{L}_{\text{train}}}$. During both validation and testing, we are only given a few trimmed support videos with corresponding long untrimmed query video. The data is divided such that $\mathcal{L}_{\text{train}} \cup \mathcal{L}_{\text{val}} \cup \mathcal{L}_{\text{test}} = \emptyset$.

\subsection{Architecture}
We propose an end-to-end network to solve the few-shot common action localization problem. A single query video and a few support videos are fed into the backbone, a C3D network~\cite{tran2015learning}, to obtain video representations. The weights of the backbone network are shared between the support and query videos. For the query video, a  proposal subnet predicts temporal segments of variable length containing potential activities~\cite{xu2017r}. Let $\mathcal{F}_Q \in \mathbb{R}^{R \times C}$ denote the feature representation of the query video for $R$ temporal proposal segments, each of dimensionality $C$. Let $\mathcal{F}_S \in \mathbb{R}^{ST \times C}$ denote the representations of the $S$ support videos, where we split each support video into $T$ fixed temporal parts. The main goal of the network is to align the support representations with the relevant query segment representation:
\begin{equation}
\mathcal{F} = \phi(\mathcal{F}_Q, \mathcal{F}_S).
\label{eq:mapping}
\end{equation}
In Equation~\ref{eq:mapping}, $\mathcal{F} \in \mathbb{R}^{R \times C}$ denotes the temporal segment representations after alignment with the support representations through $\phi$. In our common localization network, representations $\mathcal{F}$ are fed to fully-connected layers that perform a binary classification to obtain the likelihood that each proposal segment matches with the support actions, which is followed by a temporal regression to refine the activity start- and end-times for all segments.

In our network, we consider the following: \textit{i}) the representations of the support videos need to be aligned with the representations of the activity in the query video, \textit{ii}) not all support videos are equally informative, and \textit{iii}) common action localization is a support-conditioned localization task, where the activityness of different query segments should be guided by the support videos. We propose three modules, namely mutual enhancement module, progressive alignment module, and pairwise matching module to deal with these considerations.

\begin{figure*}[t]
	\centering
	\includegraphics[width=0.99\linewidth]{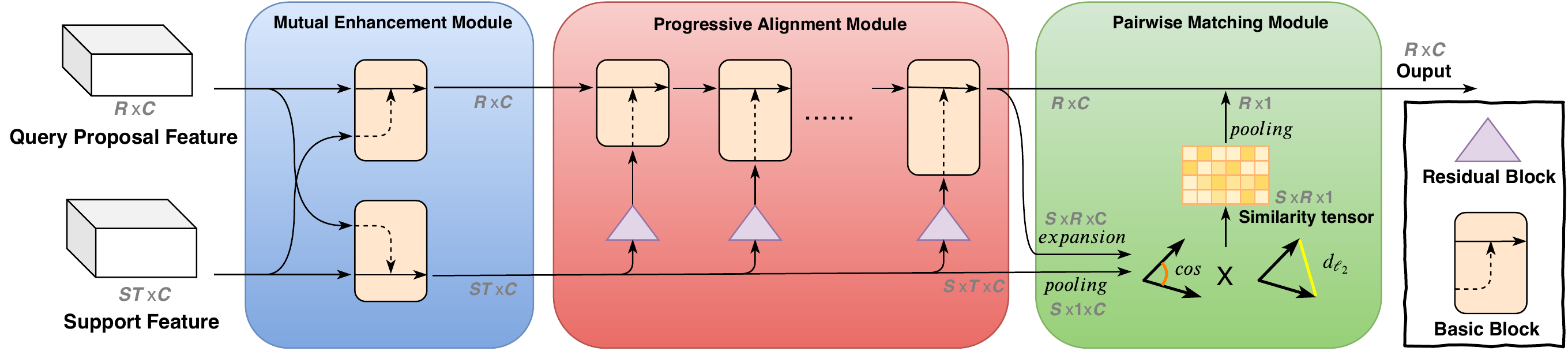}
	\caption{\textbf{Modules for aligning representations} from the support videos with the relevant query video segments.
	%
	The mutual enhancement module augments the support and query representations simultaneously through message passing.
	%
	Then, the progressive alignment module fuses the support into the query branch through recursive use of the basic block.
	%
	Finally, the pairwise matching module reweighs the fused features according to the similarity between the enhanced query segments and the enhanced support videos.
	} 
	\label{fig:modules}
\end{figure*}

\subsubsection{Mutual enhancement module.}
Building on the recent success of the transformer structure~\cite{transformer} and the non-local block~\cite{nl}, which are forms of self-attention, we propose a module which can simultaneously enhance the representations of the support and query videos from each other. The basic block for this module is given as: 
\begin{equation}
\begin{split}
m(\mathit{I}_1, \mathit{I}_2) = 
c_{1}(\textit{soft}(c_{2}(\mathit{I}_1) \times 
 c_{3}(\mathit{I}_{2}^{T})) \times c_{4}(\mathit{I}_{2})) + \mathit{I}_1,
\label{equ:non_local}
\end{split}
\end{equation}
where \(c_{1},c_{2},c_{3},c_{4}\) are fully-connected layers, $\textit{soft}$ denotes the softmax activation, and \( \times\) denotes matrix multiplication. $I_1$ and $I_2$ denote the two inputs. A detailed overview and illustration of the basic block is provided in the supplementary materials. Based on the basic block, we design a mutual enhancement module to learn mutually-enforced representations, both for query proposals and support videos, as shown in Figure~\ref{fig:modules}. The mutual enhancement module has two streams $m_{s \rightarrow q}$, $m_{q \rightarrow s}$ that are responsible for enhancing query proposals and support videos respectively. The inputs to the mutual enhancement module, $\mathcal{F}_Q$ and $\mathcal{F}_S$, will be enhanced by each other:
\begin{equation}
    m_{s \rightarrow q} = m(\mathcal{F}_Q, \mathcal{F}_S),
\end{equation}
\begin{equation}
    m_{q \rightarrow s} = m(\mathcal{F}_S, \mathcal{F}_Q).
\end{equation}

\subsubsection{Progressive alignment module.}
We also propose a progressive alignment module to achieve a better fusion of representations from query proposals and support videos. The idea behind this module is to reuse the basic block from the mutual enhancement module to integrate the support branch into the query branch. Inspired by the successful application of residual learning~\cite{he2016deep,hu2018squeeze}, we employ a residual block to make the progressive alignment effective:
\begin{equation}
    r(\mathit{I})=\mathit{c}_1(\textit{relu}(c_{2}(\mathit{I}))) + \mathit{I},
\end{equation}
where $\mathit{c}_1$, $\mathit{c}_2$ are fully-connected layers, \textit{relu} denotes the ReLU activation. A detailed overview and illustration of the residual block is provided in the supplementary materials. We first take query proposal representations from the first module $m_{s \rightarrow q}$ as 0-depth outcome $\mathcal{P}_{0}$. On top, we adopt our basic block $m$ to integrate this outcome with $m_{q\rightarrow s}$ which has been recalibrated by our residual block $r$. We perform this operation multiple times in a recursive manner, \ie:
\begin{equation}
    \mathcal{P}_{0} = m_{s \rightarrow q},
\end{equation}
\begin{equation}
    \mathcal{P}_{k} = m(\mathcal{P}_{k-1}, \mathit{r}(m_{q \rightarrow s})),\quad k=1,2,\dots,n.
\end{equation}
Where we set $n=3$ in practice. The advantage of a progressive design is that it strengthens the integration of the support branch into the query branch as we increase the number of basic block iterations. By using the same efficient basic blocks as our first module, the computational overhead is small. An illustration of the progressive alignment module is shown in Figure~\ref{fig:modules}.

\subsubsection{Pairwise matching module.}
In common action localization, a small number of support videos is used. Intuitively, not every support video is  equally informative for the query segments. In addition, different query segments should not be treated equally either. To address these intuitions, we add a means to weigh the influence between each support video and each query segment, by introducing a pairwise matching module. 

The input for the matching module are all segments of the query video and all support videos. 
The pair-wise matching is a mapping $\text{PMM}: (\mathbb{R}^{R \times C},\mathbb{R}^{S \times T \times C})  \mapsto  \mathbb{R}^{S \times R \times 1}$. To align the two components, we first perform an expansion operation $e$ on the query segments, denoted as $e(\mathcal{P}_n) \in \mathbb{R}^{S \times R \times C}$. Then a pooling $p$ is applied over the support videos along the temporal dimension, denoted as $p(m_{q \rightarrow s}) \in \mathbb{R}^{S \times 1 \times C}$. Afterwards, we perform an auto broadcasting operation $b$ on $p(m_{q \rightarrow s})$, which can broadcast the dimension of $p(m_{q \rightarrow s})$ from $\mathbb{R}^{S \times 1 \times C}$ to $\mathbb{R}^{S \times R \times C}$ to align with the dimension of $e(\mathcal{P}_{n})$.
For query segments $\mathcal{P}_n$ and for support videos $m_{q \rightarrow s}$, their match is given by the cosine similarity ($\mathit{cos}$) and $\ell_{2}$ Euclidean distance $(\mathit{d}_{\ell_{2}})$ along the segment axis:
%
\begin{equation}
M = \mathit{cos}(\mathcal{P}_{n},m_{q \rightarrow s})=\frac{<\!e(\mathcal{P}_{n}),b(p(m_{q \rightarrow s}))\!>}{\| e(\mathcal{P}_{n})\| \cdot \|b(p(m_{q \rightarrow s}))\|},
\label{eq:m}
\end{equation}

\begin{equation}
N = \mathit{d}_{\ell_{2}}(\mathcal{P}_{n},m_{q \rightarrow s})=\|e(\mathcal{P}_{n}) - b(p(m_{q \rightarrow s}))\|.
\label{eq:n}
\end{equation}
We combine both distance measures:

\begin{equation}
\mathcal{W} = \text{PMM}(\mathcal{P}_{n},m_{q \rightarrow s}) = M \odot \sigma(-N),
\end{equation}
where $\sigma$ denotes the {Sigmoid} operation. Tensor $\mathcal{W}\in \mathbb{R}^{S \times R \times 1}$ can be interpreted as a weight tensor to achieve attention over the $R$ and $S$ dimensions.  $\mathcal{W}\left[i,j\right]$ is a scalar depicting the similarity between the $j$-th query segment representation and the $i$-th support representation. For the $j$-th query segment representation, $\mathcal{W}\left[:,j\right] \in \mathbb{R}^{S \times 1}$ corresponds to the weight for different support videos, while for the $i$-th support representation, $\mathcal{W}\left[i,:\right] \in \mathbb{R}^{R \times 1}$ resembles the weight for different query segments. In the end, we enforce the pairwise matching weight $\mathcal{W}$:
\begin{equation}
\phi(\mathcal{F}_Q,\mathcal{F}_S) =\mathcal{P}_n \odot \text{AP}(\mathcal{W}),
\label{equ:mn}
\end{equation}
where AP denotes an average pooling operation along the support dimension, in other words, AP $: \mathbb{R}^{S \times R \times 1} \mapsto \mathbb{R}^{R \times 1}$.


\subsection{Optimization}
To optimize our network on the training set, we employ both a classification loss and a temporal regression loss. Different than \eg, R-C3D~\cite{xu2017r}, our classification task is specifically dependent on the few support videos. Accordingly, the loss function is given as:
%
\begin{equation}
L = \frac{1}{N_{cls}} \sum_{i} L_{cls}(a_{i},a^{*}_{i}) +  \frac{1}{N_{reg}}\sum_{i} a^{*}_{i} L_{reg} (t_{i}, t^{*}_{i}),
\end{equation}
where $N_{cls}$ and $N_{reg}$ stand for batch size and the number of proposal segments,
while $i$ denotes the proposal segment index in a batch, $a_{i}$ is the predicted probability of the proposal segment, $a^{*}_{i}$ is the ground truth label, and $t_{i}$ represents predicted relative offset to proposals. In the context of this work, the ground truth label is class-agnostic and hence binary (foreground/background), indicating the presence of an action or not. Lastly, $t^{*}_{i}$ represents the coordinate transformation of ground truth segments to proposals. 

The above loss function is applied on two parts: the support-agnostic part and the support-conditioned part. All losses for the two parts are optimized jointly. In the support-agnostic part, the foreground/background classification loss $L_{cls}$ predicts whether the proposal contains an activity, or not, and the regression loss $L_{reg}$ optimizes the relative displacement between proposals and ground truths.  For the support-conditioned part, the loss $L_{cls}$ predicts whether the proposal has the same common action as the one among the few support videos.
%
The regression loss $L_{reg}$ optimizes the relative displacement between activities and ground truths. We note explicitly that this is done for the training set only.

During inference, the proposal subnet generates proposals for the query video. The proposals are refined by Non-Maximum Suppression (NMS) with a threshold of 0.7. Then the selected proposals are fused with the support videos through the mutual enhancement, progressive alignment, and pairwise matching modules. The obtained representation is fed to the classification subnet to again perform binary classification and the boundaries of the predicted proposals are further refined by the regression layer. Finally, we conduct NMS based on the confidence scores of the refined proposals to remove redundant ones, and the threshold in NMS is set a little bit smaller than the overlap threshold $\theta$ in evaluation ($\theta=0.1$ in this paper). 

\subsubsection{Optimizing for long videos.}
The longer the untrimmed query video, the larger the need for common localization, as manual searching for the activity becomes problematic. In our setup, the length of the input video is set to 768 frames to fit the GPU memory. When the query video is longer than 768 frames, we employ multi-scale segment generation~\cite{shou2016temporal}. We apply temporal sliding windows of 256, 512, and 768 frames with 75\% overlap. Consequently, we generate a set of candidates $\Phi=\left\{(s_{h},\psi_{h},\psi_{h}^\prime)\right\}_{h=1}^H$ as input for the proposal subnet, where H is the total number of sliding windows, and $\psi_{h}$ and $\psi_{h}^\prime$ are the starting time and ending time of the $h$-th segment $s_{h}$. All refined proposals of all candidate segments together go through the NMS to remove redundant proposals. 

%% file: 3_experiment.tex
\begin{table}[t] 
\caption{\textbf{Overview of the common (multi-)instance datasets.} The common instance datasets contain a single target action per video, while the common multi-instance datasets contain more frames and more actions per video, adding to the challenge of few-shot common action localization.}
\label{tab:dataset}
\centering
 \resizebox{0.89\columnwidth}{!}{
\begin{tabular}{lrrrr}
\toprule
 & \multicolumn{2}{c}{\textbf{\qquad  Common instance \qquad}} & \multicolumn{2}{c}{\textbf{\qquad Common multi-instance}}\\
 & \quad ActivityNet & Thumos &\qquad\quad ActivityNet & Thumos\\
\midrule
\rowcolor{mygray}
\textbf{Video statistics} & & & & \\
number of instances & 1 & 1 &1.6 &14.3\\
number of frames &266.9 &284.6&444.5&5764.2\\
length (sec) &89.0&11.4 & 148.2&230.6  \\
number of train videos &10035 & 3580&6747&1665  \\
number of val+test videos &2483 & 775&1545&323  \\
\midrule
\rowcolor{mygray}
\textbf{Class statistics} & & & & \\
number of train actions &160 & 16&160&16  \\
number of val+test actions &40 & 4&40&4  \\
\bottomrule
\end{tabular}
}
\end{table}
\section{Experimental setup}

\subsection{Datasets}
Existing video datasets are usually created for classification~\cite{kay2017kinetics,idrees2017thumos}, temporal localization~\cite{caba2015activitynet}, captioning~\cite{chen2011collecting}, or summarization~\cite{gygli2014creating}. To evaluate few-shot common action localization, we have revised two existing datasets, namely ActivityNet1.3~\cite{caba2015activitynet} and Thumos14~\cite{idrees2017thumos}. Both datasets come with temporal annotations suitable for our evaluation. We consider both common instance and common multi-instance, where the latter deals with query videos containing multiple instances of the same action.

\textbf{Common instance.} 
For the revision of ActivityNet1.3, we follow the organization of Feng \etal~\cite{feng2018video}. We divide videos that contain multiple actions into independent videos, with every newly generated video consisting of just one action and background. Next we discard videos longer than 768 frames. We split the remaining videos into three subsets, divided by action classes. We randomly select 80\% of the classes for training, 10\% of the classes for validation, and the remaining 10\% of the classes for testing. Besides ActivityNet, we also revise the Thumos dataset using the same protocol.

\begin{SCtable}
	\caption{\textbf{Module evaluation} on ActivityNet and Thumos in the common instance setting. All three modules have a positive mAP effect on the localization performance with only a slight increase in parameters.}
	\label{tab:nl vs similarity all}
	\centering
	\resizebox{0.55\columnwidth}{!}{%
	\begin{tabular}{ccccccc}
	\toprule
	&&&\multicolumn{2}{c}{\textbf{ActivityNet}}&\multicolumn{2}{c}{\textbf{Thumos}}\\
	MEM & PAM & PMM & one-shot & five-shot & one-shot & five-shot\\
	\midrule
	& & & {42.4} & {42.5} & {37.5}  & {38.4}\\
    & $\checkmark$ & & {49.7} & {52.0} & {42.3}  & {44.5}\\
    & $\checkmark$ & $\checkmark$ & {51.3} & {53.6} & {44.8}  & {46.0}\\
	$\checkmark$ & $\checkmark$ & & {52.5} & {55.3} & {47.6}  & {49.6}\\
	\rowcolor{mygray}
	$\checkmark$ & $\checkmark$ & $\checkmark$ & \textbf{53.1} & \textbf{56.5} & \textbf{48.7}  & \textbf{51.9}\\
	\bottomrule
    \end{tabular}%
	}
\end{SCtable}

\begin{figure}[tb!]
	\centering
	\begin{subfigure}{0.49\textwidth}
	\includegraphics[width=\textwidth]{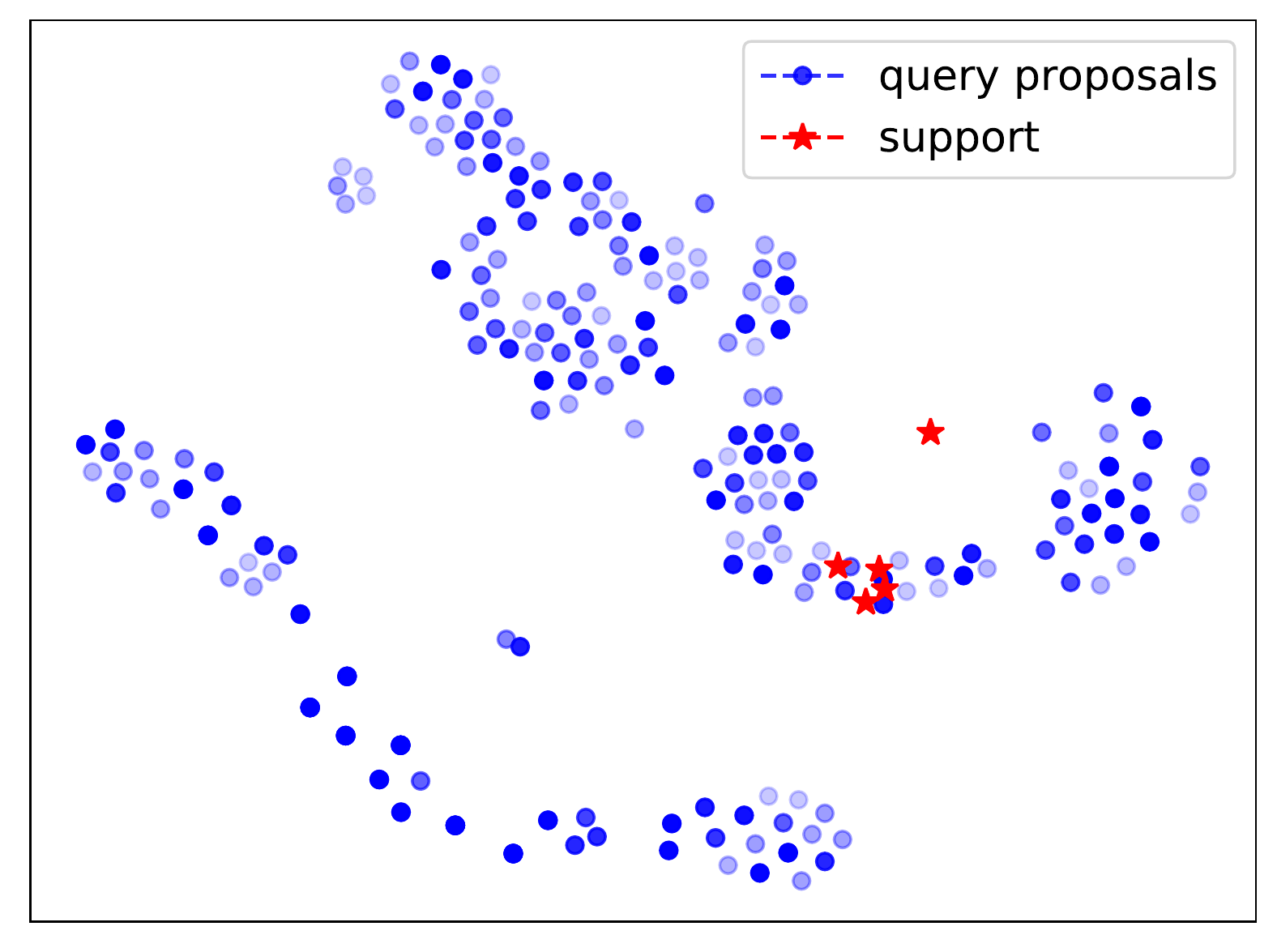}
	\caption{Without our modules.}
	\label{fig:tsne_before}
	\end{subfigure}
	\begin{subfigure}{0.49\textwidth}
	\includegraphics[width=\textwidth]{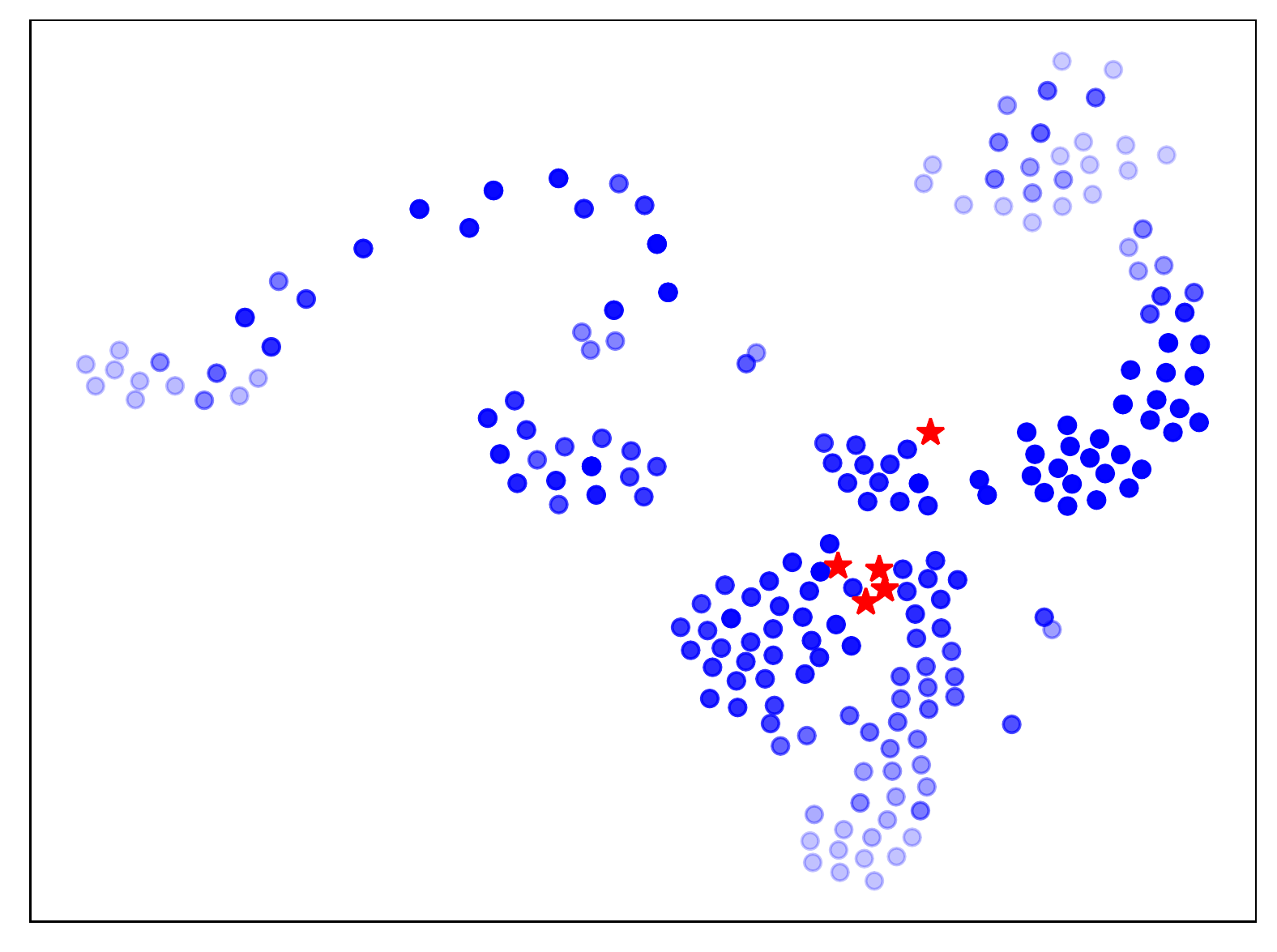}
	\caption{With our modules.}
	\label{fig:tsne_after}
	\end{subfigure}
	\caption{\textbf{Module evaluation} by t-SNE visualization of support and query representations. Colors of query proposals indicate their overlap with the ground truth action, the darker the better. Without our modules (left), both relevant and irrelevant query proposals are near the support videos. Afterwards (right), only relevant proposals remain close to the support videos, highlighting the effectiveness of our modules for localizing common actions among a few videos.}
	\label{fig:tsne}
\end{figure}

\textbf{Common multi-instance.}
Query videos in real applications are usually unconstrained and contain multiple action segments. Therefore, we also split the original videos of ActivityNet1.3 and Thumos14 into three subsets according to their action classes without any other video preprocessing. As a result, we obtain long query videos with multiple action instances. The support videos are still trimmed action videos. 

During training, the support videos and query video are randomly paired, while the pairs are fixed for validation and testing. The differences between the common instance and common multi-instance video datasets are highlighted in Table~\ref{tab:dataset}.

\subsection{Experimental details}
We use PyTorch~\cite{pytorch} for implementation. Our network is trained with Adam~\cite{kingma2014adam} with a learning rate of 1e-5 on one Nvidia GTX 1080TI. We use 40k training iterations and learning rate is decayed to 1e-6 after 25k iterations. 
To be consistent with the training process of our baselines~\cite{feng2018video,zhang2019localizing}, we use the same C3D backbone~\cite{tran2015learning}. The backbone is pre-trained on Sports-1M~\cite{karpathy2014large} and is fine-tuned with a class-agnostic proposal loss on the training videos for each dataset.
%
The batch size is set to 1. The proposal score threshold is set as 0.7. The proposal number after NMS is 128 in training and 300 in validation and testing.

\subsection{Evaluation}
Following~\cite{shou2016temporal,feng2018video}, we measure the localization performance using (mean) Average Precision. A prediction is correct when it has the correct foreground/background prediction and has a ground truth overlap larger than the overlap threshold. The overlap is set to 0.5 unless specified otherwise. 

\section{Experimental results}
\subsection{Ablation study}


\textbf{Module evaluation.} We  evaluate the effect of the mutual enhancement module (MEM), the progressive alignment module (PAM), and the pairwise matching module (PMM) for our task on the common instance datasets. We report results using one and five support videos in Table~\ref{tab:nl vs similarity all}. To validate the effectiveness of our modules, we compare to our baseline system without any modules. Here the support representations are averaged and added to the query representations. We observe that the progressive alignment module increases over the baseline considerably, showing its efficacy. Adding the pairwise matching on top of the progressive alignment or using the mutual enhancement before the progressive alignment further benefits few-shot common action localization. Combining all three modules works best. 

To get insight into the workings of our modules for common action localization, we have analysed the feature distribution before and after the use of our modules. In Figure~\ref{fig:tsne}, we show the t-SNE embedding~\cite{maaten2008visualizing} before and after we align the five support videos with the 300 proposals in one query video. We observe that after the use of our modules, the proposals with high overlap are closer to the support videos, indicating our ability to properly distill the correct action locations using only a few support videos. Irrelevant proposals are pushed away from the support videos, which results in a more relevant selection of action locations.

\begin{SCtable}
	\caption{\textbf{Influence of noisy support videos} on common-instance ActivityNet for the five-shot setting. The result shows that our approach is robust to the inclusion of noisy support videos, whether they come from the same or different classes.}
	\label{tab:vedio level noise}
	\centering
	\resizebox{0.50\columnwidth}{!}{
	\begin{tabular}{lc}
			\toprule
			No noise & 56.5\\
			\midrule
			1 noisy support video & 53.5\\
			2 noisy support videos of different class & 51.9\\
			2 noisy support videos of same class & 50.6\\
			\bottomrule
		\end{tabular}%
		}
\end{SCtable}

\textbf{Few-shot evaluation.}
Our common action localization is optimized to work with multiple examples as support. To show this capability, we have measured the effect of gradually increasing the number of support videos, we found that the mAP gradually increases as we enlarge the number of support videos from one to six on common-instance ActivityNet. We obtain an mAP of 53.1 (one shot), 53.8 (two shots), 54.9 (three shots), 55.4 (four shots), 56.5 (five shots), 56.8 (six shots).
The results show that our approach obtains high accuracy with only a few support videos. Using more than one support video is beneficial for common action localization in our approach, showing that we indeed learn from using more than one support video. Results stagnate when using more than six examples.

\begin{figure}[tb!]
	\centering
	\begin{subfigure}{0.49\textwidth}
	\includegraphics[width=\textwidth]{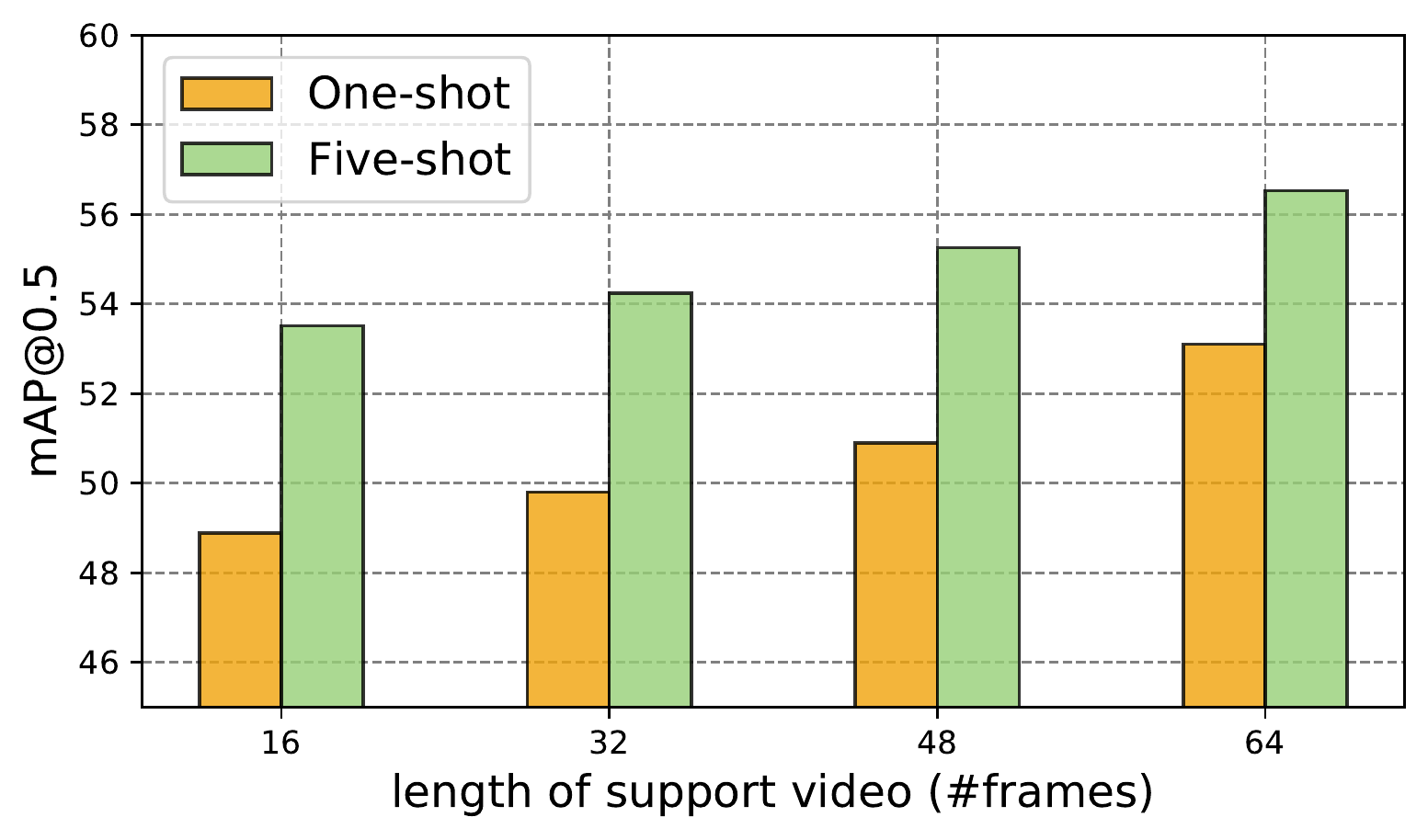}
	\caption{Effect of support video length.}
	\label{fig:support_video_length}
	\end{subfigure}
	\begin{subfigure}{0.49\textwidth}
	\includegraphics[width=\textwidth]{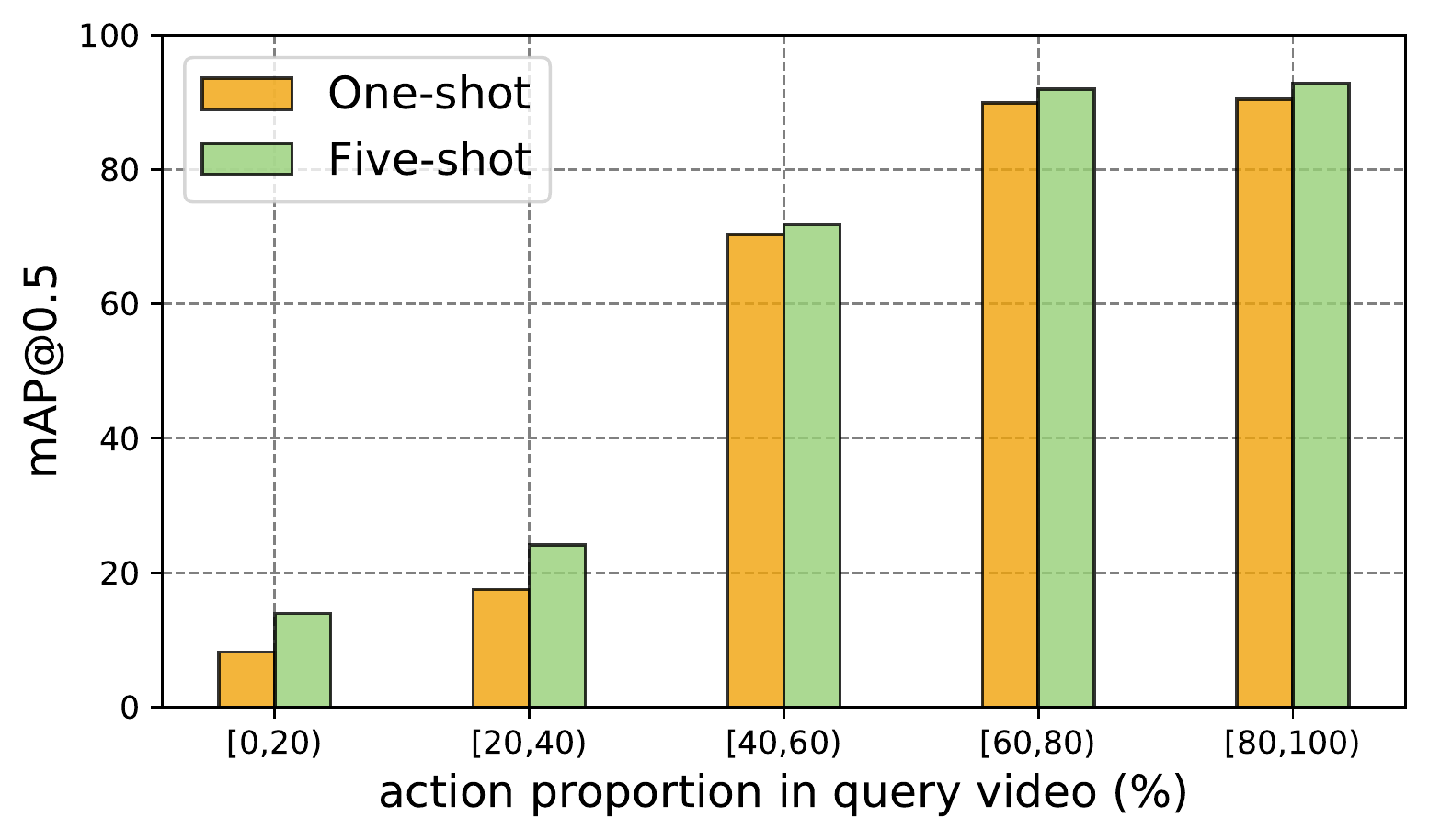}
	\caption{Effect of action ratio in query video.}
	\label{fig:result_ratio}
	\end{subfigure}
	\caption{\textbf{Ablation studies} on the length of the support videos and the action proportion in the query video. Both studies are on common-instance ActivityNet. Left: The longer the support videos, the better we perform, as we can distill more knowledge from the limited provided supervision. Right: High scores can be obtained when the common action is dominant, localization of short actions in long videos remains challenging.}
\end{figure}

\textbf{Effect of support video length.}
We ablate the effect of the length of the support videos on the localization performance in Figure~\ref{fig:support_video_length}. We sample 16, 32, 48 and 64 frames for each support video respectively. We find that the result gradually increases with longer support videos, which indicates that temporal information in the support videos is beneficial to our modules for common action localization.

\textbf{Influence of action proportion in query video.} Figure~\ref{fig:result_ratio} shows that for query videos with a dominant action, we can obtain high scores. An open challenge remains localizing very short actions in very long videos.

\begin{figure*}[t]
\centering
\includegraphics[width=1.0\linewidth]{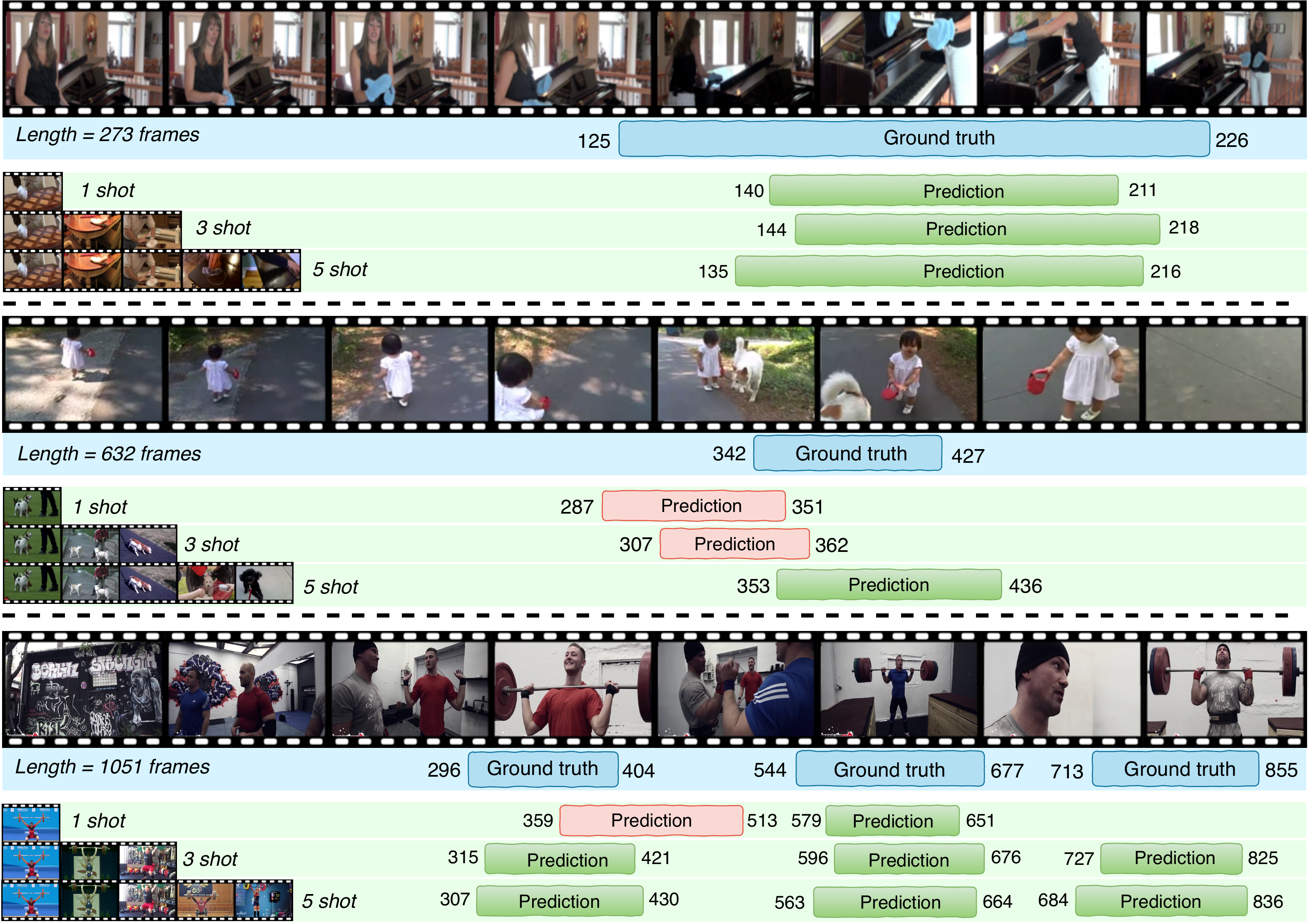}
\caption{\textbf{Qualitative result} of predictions by our approach under 1-shot, 3-shot and 5-shot settings. Correct predictions with an overlap larger than 0.5 are marked in {\color{green}green}, and incorrect predictions are marked in {\color{red}red}. The length and start-end boundary of segment are indicated in frame numbers.}
\label{fig:qualitative result 1}
\end{figure*}


\textbf{Influence of noisy support videos.} To test the robustness of our approach, we have investigated the effect of including noisy support videos in the five-shot setting. The results are shown in Table~\ref{tab:vedio level noise}. 
When one out of five support videos contains the wrong action, the performance drops only 3\% from 56.5 to 53.5. 
The performance drop remains marginal when replacing two of the five support videos with noisy videos.
When two noisy support videos are from the same class, the drop is larger, which is to be expected, as this creates a stronger bias towards a distractor class. Overall, we find that our approach is robust to noise for common action localization.


\textbf{Qualitative results.} To visualize the result of our method, we show three cases in Figure~\ref{fig:qualitative result 1}. For the first example, we can find the common action location from one support video. Adding more support videos provides further context, resulting in a better fit. For the second one, our method can recover the correct prediction only when five support videos are used. As shown in the third case, our method can also handle the multi-instance scenario. We show a query video with three instances. With only one support video, we miss one instance and have low overlap with another. When more support videos are added, we can recover both misses.

\subsection{Comparisons with others}
To evaluate the effectiveness of our proposed approach for common action localization, we perform three comparative evaluations. 


\begin{SCtable}
\caption{\textbf{One-shot comparison on common instance ActivityNet}. Results marked with * obtained with author provided code. In both settings, our approach is preferred across all overlaps, highlighting its effectiveness.}
\label{tab:activitynet}
\centering
\resizebox{0.5\columnwidth}{!}{%
\begin{tabular}{lcccccc}
\toprule
 & \multicolumn{6}{c}{\textbf{Overlap threshold}}\\
& 0.5 & 0.6 & 0.7 & 0.8 & 0.9 & 0.5:0.9\\
\midrule
\rowcolor{mygray}
\textbf{Common instance} & & & & & & \\
Hu \etal~\cite{silco2019} * &{41.0}&{33.0}&{27.1}&{15.9}&{6.8}&{24.8}\\
Feng \etal~\cite{feng2018video} &43.5&35.1&27.3&16.2&6.5&25.7 \\
\textit{This paper} &\textbf{53.1}&\textbf{40.9}&\textbf{29.8}&\textbf{18.2}&\textbf{8.4}&\textbf{29.5}\\
\midrule
\rowcolor{mygray}
\textbf{Common multi-instance} & & & & & & \\
Hu~\etal~\cite{silco2019} * & 29.6 & 23.2 & 12.7 & 7.4 & 3.1 & 15.2 \\
Feng~\etal~\cite{feng2018video} * & 31.4 & 25.5 & 16.1 & 8.9 & 3.2 & 17.0 \\
\textit{This paper} &\textbf{42.1}&\textbf{36.0}&\textbf{18.5}&\textbf{11.1}&\textbf{7.0}&\textbf{22.9}\\
\bottomrule
\end{tabular}
}
\end{SCtable}

\begin{figure}[t]
	\centering
	\includegraphics[width=1.0\columnwidth]{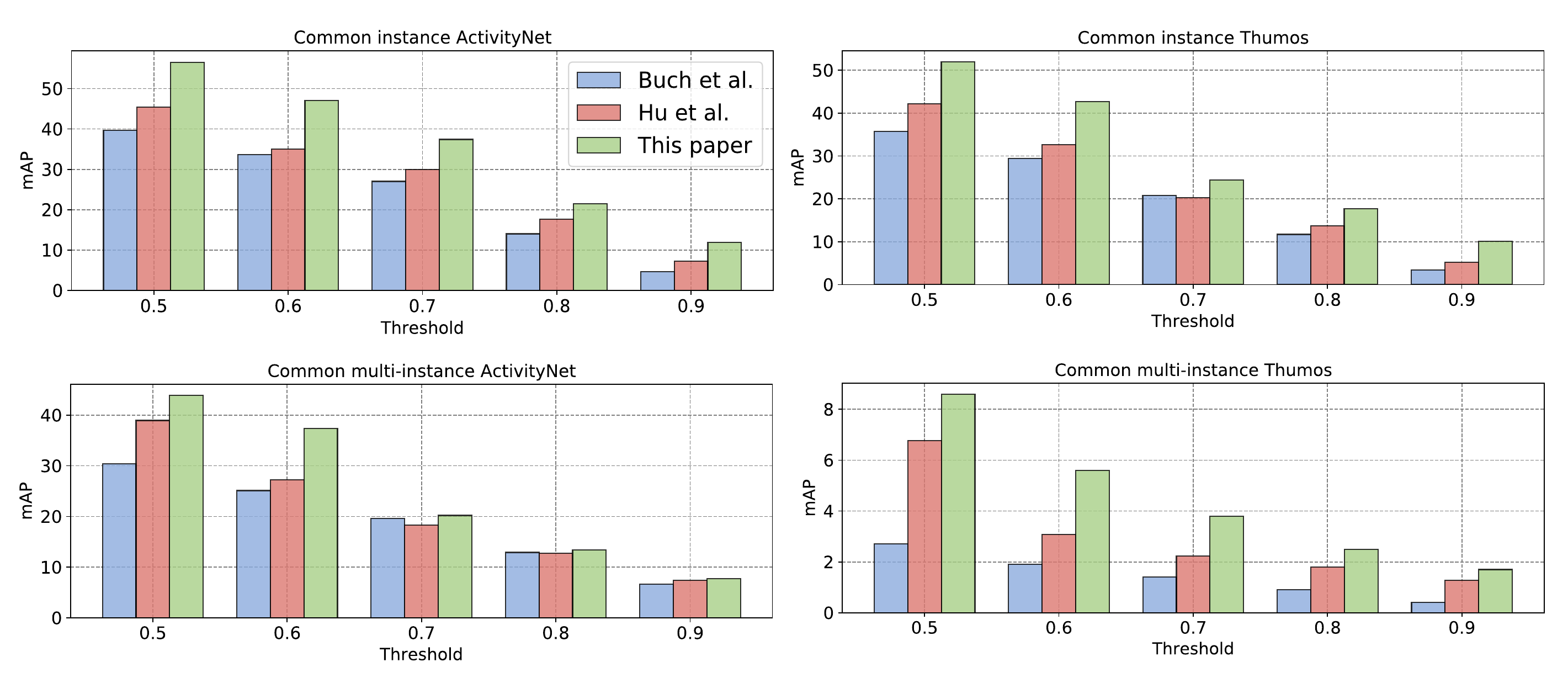}
	\caption{\textbf{Five-shot comparison.} We evaluate our method as well as modified versions of Hu~\etal~\cite{silco2019} and Buch \etal~\cite{buch2017sst} on all common instance and multi-instance datasets,  we  obtain  favourable  results. Detailed numerical results are provided in supplementary file to facilitate the comparison for the follow-up works. Best viewed in color.}
	\label{fig:fourtable}
\end{figure}

\textbf{One-shot comparison.}
For the one-shot evaluation, we compare to the one-shot video re-localization of Feng \etal~\cite{feng2018video} and to Hu \etal~\cite{silco2019}, which focuses on few-shot common object detection.
We evaluate on the same setting as Feng \etal~\cite{feng2018video}, namely the revised ActivityNet dataset using the one-shot setting (common instance). Note that we both use the C3D base network. To evaluate the image-based approach of Hu \etal~\cite{silco2019}, we use their proposed similarity module on the temporal video proposals, rather than spatial proposals based on author provided code~\cite{silco2019}. The results in Table~\ref{tab:activitynet} show that across all overlap thresholds, our approach is preferred. At an overlap threshold of 0.5, we obtain an mAP of 53.1 compared to 41.0 for~\cite{silco2019} and 43.5 for~\cite{feng2018video}.
It is of interest to note that without our three modules, we obtain only 42.4 (Table~\ref{tab:nl vs similarity all}). This demonstrates that a different training setup or a different model architecture by itself does not benefit common action localization. We attribute our improvement to the better alignment between the support and query representations as a result of our three modules. 
Next to a comparison on the common instance dataset, we also perform the same experiment on the longer multi-instance ActivityNet variant. In this more challenging setting, our approach again outperforms the baselines.  We note that we are not restricted to the one-shot setting, where the baseline by Feng \etal~\cite{feng2018video} is.

\begin{SCtable}
\caption{\textbf{Localization from images} on the common instance datasets. 
Our method generalizes beyond videos as support input and outperforms Zhang \etal~\cite{zhang2019localizing}}
\label{tab:image_query}
\centering
\resizebox{0.50\columnwidth}{!}{%
\begin{tabular}{lcccc}
\toprule
& \multicolumn{2}{c}{\textbf{ActivityNet}} & \multicolumn{2}{c}{\textbf{Thumos}}\\
& one-shot & five-shot& one-shot & five-shot \\
\midrule
Zhang \etal &45.2&48.5&36.9&38.9\\
\rowcolor{mygray}
\textit{This paper} &\textbf{49.2}&\textbf{52.8}&\textbf{43.0}&\textbf{45.6}\\
\bottomrule
\end{tabular}
}
\end{SCtable}

\textbf{Five-shot comparison.} Second, we evaluate the performance of our approach on all datasets in the five-shot setting. We compare to a modified version of SST by Buch \etal~\cite{buch2017sst}. We add a fusion layer on top of the original GRU networks in SST to incorporate the support feature, and then choose the proposal with the largest confidence score. SST is used as baseline, because the approach of Feng \etal~\cite{feng2018video} cannot handle more than one support video. We also include another comparison to Hu \etal~\cite{silco2019}. This time also using their feature reweighting module. The results are shown in Figure~\ref{fig:fourtable}. We observe that our method performs favorably compared to the two baselines on all datasets, reaffirming the effectiveness of our method. 
Also note that even when our support videos are noisy (Table~\ref{tab:vedio level noise}), we are still better than the baselines without any noise
based on Buch~\etal~\cite{buch2017sst} and Hu~\etal~\cite{silco2019} (39.7 and 45.4 for a threshold of 0.5 on common instance ActivityNet).
The large amount of distractor actions in the long videos of common multi-instance Thumos results in lower overall scores, indicating that common action localization is far from a solved problem.


\textbf{Localization from images.}
Next to using videos, we can also perform common action localization using images as support. This provides a challenging setting, since any temporal information is lost. We perform localization from support images by inflating the images to create static support videos. We perform a common action localization on common instance ActivityNet and Thumos. We compare to the recent approach of Zhang \etal~\cite{zhang2019localizing}, which focuses on video retrieval from images. Results in Table~\ref{tab:image_query} show we obtain favourable results on both datasets, even though our approach is not designed for this setting. 


%% file: 4_conclusion.tex
\section{Conclusion}
In this paper we consider action localization in a query video given a few trimmed support videos that contain a common action, without specifying the label of the action. To tackle this challenging problem, we introduce a new network architecture along with three modules optimized for temporal alignment. The first module focuses on enhancing the representations of the query and support representation simultaneously. The second module progressively integrates the representations of the support branch into the query branch, to distill the common action in the query video. The third module weighs the different support videos to deal with non-informative support examples. Experiments on reorganizations of ActivityNet and Thumos dataset, both with settings containing a single and multiple action instances per video, show that our approach can robustly localize the action which is common amongst support videos in both standard and long untrimmed query videos.

%% file: 6_supp.tex
\appendix

\section{Dataset}

\textbf{Detail of class statistics.} We list the class names for the train, validation and test sets of the {common instance} and {multi-instance ActivityNet} in Table~\ref{tab:activitynet-label}, and for the {common instance} and {multi-instance Thumos} in Table~\ref{tab:thumos-label}.

\section{Method}

\begin{figure}[tb!]
	\centering
	\begin{subfigure}{0.59\textwidth}
	\includegraphics[width=\textwidth]{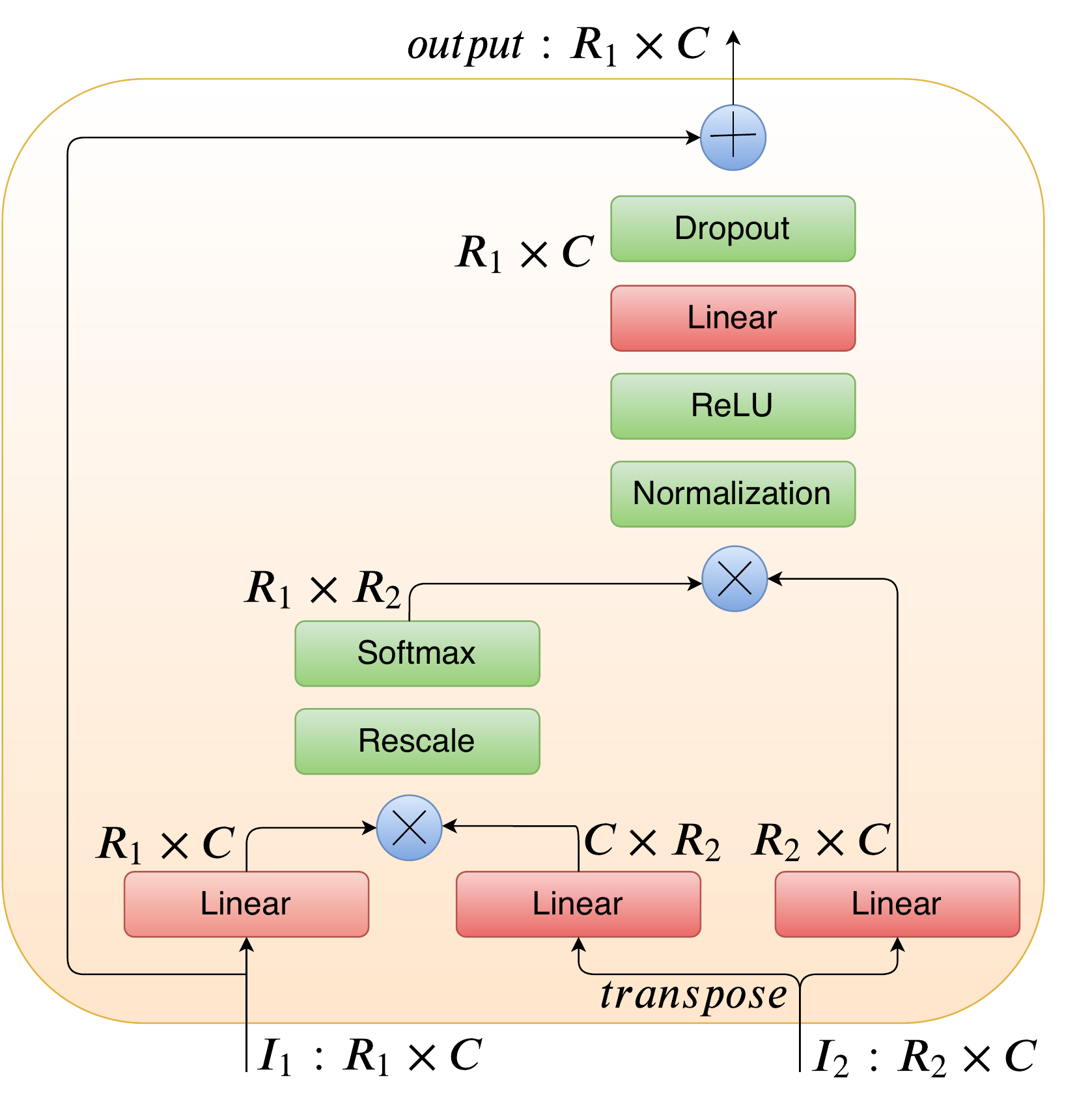}
	\caption{Scheme of basic block.}
	\label{fig:basic_block}
	\end{subfigure}
	\begin{subfigure}{0.36\textwidth}
	\includegraphics[width=\textwidth]{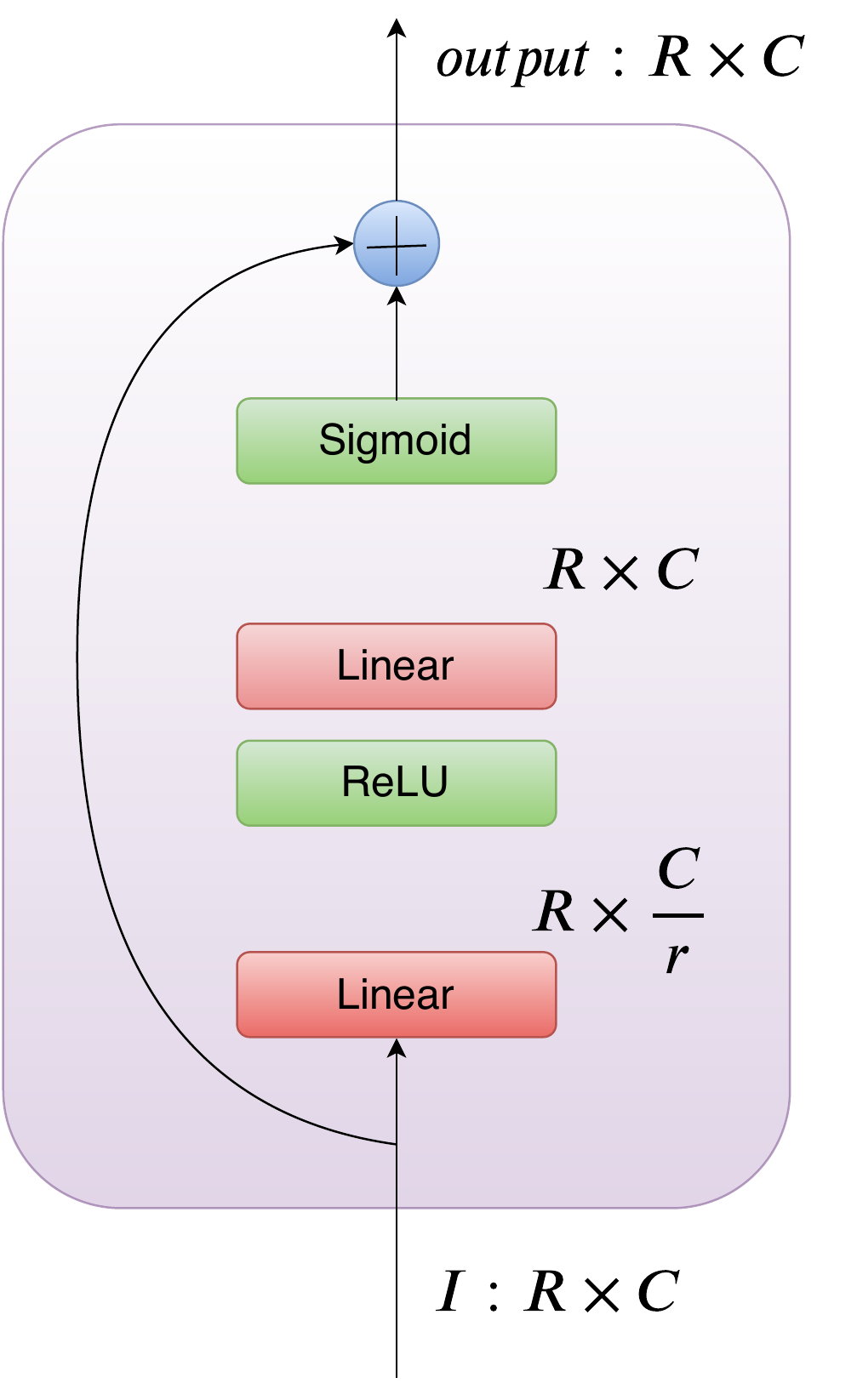}
	\caption{Scheme of residual block.}
	\label{fig:residual_block}
	\end{subfigure}
	\caption{\textbf{Overview of our blocks.} $I_1$, $I_2$ denote the inputs of our basic block (left) and $I$ denotes the input of our residual block (right). $\otimes$ denotes matrix multiplication and $\oplus$ is  element-wise sum. $r$ is set to 4.}
	\label{fig:tsne}
\end{figure}

\textbf{Overview of the basic block and the residual block.} The structure of the basic block is illustrated in Figure~\ref{fig:basic_block}. The main idea of the basic block is to align the features $I_2$ to the features $I_1$. Figure~\ref{fig:residual_block} depicts the schema of the residual block. Here, the residual block performs recalibration on the input features $I$.

\section{Results}

\textbf{Numerical results of five-shot comparison}. Table~\ref{tab:five_shot_comparison} displays the numerical results of the five-shot comparison between our method and modified versions of Buch~\etal~\cite{buch2017sst} and Hu~\etal~\cite{silco2019} on both the common instance and multi-instance datasets. The numerical results are provided here to facilitate the comparison for the follow-up works.

\begin{table}
\caption{\textbf{Numerical results of five-shot comparison.} 
}
\label{tab:five_shot_comparison}
\centering
\resizebox{0.99\columnwidth}{!}{%
\setlength\tabcolsep{4pt}
\begin{tabular}{lcccccc|cccccc}
\toprule
 &\multicolumn{6}{c}{\textbf{ActivityNet}}&\multicolumn{6}{c}{\textbf{Thumos}}\\
 & 0.5 & 0.6 & 0.7 & 0.8 & 0.9 & 0.5:0.9 & 0.5 & 0.6 & 0.7 & 0.8 & 0.9 & 0.5:0.9\\
\midrule
\rowcolor{mygray}
\multicolumn{13}{l}{\textbf{Common instance}}\\
Buch~\etal & 39.7 & 33.6 & 27.0 & 14.0 & 4.6 & 23.3 & 35.7 & 29.4 & 20.8 & 11.7 & 3.4 & 20.2\\
Hu~\etal & 45.4 & 35.0 & 29.9 & 17.6 & 5.2 & 27.0 & 42.2 & 32.6 & 20.3 & 13.7 & 5.2 & 22.8\\
\textit{This paper} &\textbf{56.5}&\textbf{47.0}&\textbf{37.4}&\textbf{21.5}&\textbf{11.9}&\textbf{34.9}&\textbf{51.9}&\textbf{42.7}&\textbf{24.4}&\textbf{17.7}&\textbf{10.1}&\textbf{29.3}\\
\midrule
\rowcolor{mygray}
\multicolumn{13}{l}{\textbf{Common multi-instance}}\\
Buch~\etal & 30.4 & 25.1 & 19.6 & 12.9 & 6.6 & 18.9& 2.7 & 1.9 & 1.4 & 0.9 & 0.4 & 1.5 \\
Hu~\etal & 38.9 & 27.2 & 18.3 & 12.7 & 7.3 & 20.9 & 6.8 & 3.1 & 2.2 & 1.8 & 1.3 & 3.1\\
\textit{This paper} &\textbf{43.9}&\textbf{37.4}&\textbf{20.2}&\textbf{13.4}&\textbf{7.7}&\textbf{24.5}&\textbf{8.6}&\textbf{5.6}&\textbf{3.8}&\textbf{2.5}&\textbf{1.7}&\textbf{4.4}\\
\bottomrule
\end{tabular}
}
\end{table}

\textbf{Effect of depth in the progressive alignment module.} $n$ in Equation~{\color{red}7} denotes the depth of the progressive alignment module which is set to 3 in practice. We ablate the effect of the depth of the progressive alignment module in Table~\ref{tab:depth} under the five-shot setting on the common instance ActivityNet dataset. From depth 1 to 3, the mAP improves steadily with only a slight increase in parameters. Results stagnate when the depth is higher than 3.

\begin{table}
\caption{\textbf{Ablation study on the depth of the progressive alignment module} under five-shot setting on common instance ActivityNet. From depth 1 to 3, the mAP improves steadily with a slight increase in parameters. Results stagnate when the depth is more than 3.}
\label{tab:depth}
\centering
\resizebox{0.6\columnwidth}{!}{%
\setlength\tabcolsep{4pt}
\begin{tabular}{lccccc}
\toprule
 & \multicolumn{5}{c}{\textbf{ $n$-depth}}\\
 & 1 & 2 & 3 & 4 & 5  \\
\midrule
\#param & 47.2M & 48.2M & 49.3M & 50.3M & 51.4M  \\
{mAP} &\textbf{53.9}&\textbf{55.6}&\textbf{56.5}&\textbf{56.7}&\textbf{56.4}\\
\bottomrule
\end{tabular}
}
\end{table}

\begin{table*}
    \caption{\textbf{Subset labels on common (multi-)instance ActivityNet.}  }
    \label{tab:activitynet-label}
    \centering
    \begin{tabularx}{\textwidth}{X}
        \toprule
        \textit{\textbf{Training:}} {Fun sliding down, Beer pong, Getting a piercing, Shoveling snow, Kneeling, Tumbling, Playing water polo, Washing dishes, Blowing leaves, Playing congas, Making a lemonade, Playing kickball, Removing ice from car, Playing racquetball, Swimming, Playing bagpipes, Painting, Assembling bicycle, Playing violin, Surfing, Making a sandwich, Welding, Hopscotch, Gargling mouthwash, Baking cookies, Braiding hair, Capoeira, Slacklining, Plastering, Changing car wheel, Chopping wood, Removing curlers, Horseback riding, Smoking hookah, Doing a powerbomb, Playing ten pins, Getting a haircut, Playing beach volleyball, Making a cake, Clean and jerk, Trimming branches or hedges, Drum corps, Windsurfing, Kite flying, Using parallel bars, Doing kickboxing, Cleaning shoes, Playing field hockey, Playing squash, Rollerblading, Playing drums, Playing rubik cube, Sharpening knives, Zumba, Raking leaves, Bathing dog, Tug of war, Ping-pong, Using the balance beam, Playing lacrosse, Scuba diving, Preparing pasta, Brushing teeth, Playing badminton, Mixing drinks, Discus throw, Playing ice hockey, Doing crunches, Wrapping presents, Hand washing clothes, Rock climbing, Cutting the grass, Wakeboarding, Futsal, Playing piano, Baton twirling, Mooping floor, Triple jump, Longboarding, Polishing shoes, Doing motocross, Arm wrestling, Doing fencing, Hammer throw, Shot put, Playing pool, Blow-drying hair, Cricket, Spinning, Running a marathon, Table soccer, Playing flauta, Ice fishing, Tai chi, Archery, Shaving, Using the monkey bar, Layup drill in basketball, Spread mulch, Skateboarding, Canoeing, Mowing the lawn, Beach soccer, Hanging wallpaper, Tango, Disc dog, Powerbocking, Getting a tattoo, Doing nails, Snowboarding, Putting on shoes, Clipping cat claws, Snow tubing, River tubing, Putting on makeup, Decorating the Christmas tree, Fixing bicycle, Hitting a pinata, High jump, Doing karate, Kayaking, Grooming dog, Bungee jumping, Washing hands, Painting fence, Doing step aerobics, Installing carpet, Playing saxophone, Long jump, Javelin throw, Playing accordion, Smoking a cigarette, Belly dance, Playing polo, Throwing darts, Roof shingle removal, Tennis serve with ball bouncing, Skiing, Peeling potatoes, Elliptical trainer, Building sandcastles, Drinking beer, Rock-paper-scissors, Using the pommel horse, Croquet, Laying tile, Cleaning windows, Fixing the roof, Springboard diving, Waterskiing, Using uneven bars, Having an ice cream, Sailing, Washing face, Knitting, Bullfighting, Applying sunscreen, Painting furniture, Grooming horse, Carving jack-o-lanterns}\\
        \midrule
        \textit{\textbf{Validation:}} Swinging at the playground, Dodgeball, Ballet, Playing harmonica, Paintball, Cumbia, Rafting, Hula hoop, Cheerleading, Vacuuming floor, Playing blackjack, Waxing skis, Curling, Using the rowing machine, Ironing clothes, Playing guitarra, Sumo, Putting in contact lenses, Brushing hair, Volleyball\\
        \midrule
        \textit{\textbf{Testing:}} Hurling, Polishing forniture, BMX, Riding bumper cars, Starting a campfire, Walking the dog, Preparing salad, Plataform diving, Breakdancing, Camel ride, Hand car wash, Making an omelette, Shuffleboard, Calf roping, Shaving legs, Snatch, Cleaning sink, Rope skipping, Drinking coffee, Pole vault\\
        \bottomrule
    \end{tabularx}
\end{table*}

\begin{table*}
    \caption{\textbf{Subset labels on common (multi-)instance Thumos.} }
    \label{tab:thumos-label}
    \centering
    \resizebox{\columnwidth}{!}{%
    \begin{tabularx}{\textwidth}{X}
        \toprule
        \textit{\textbf{Training:}} BaseballPitch, BasketballDunk, Billiards, CleanAndJerk, CliffDiving, CricketBowling, CricketShot, Diving, FrisbeeCatch, GolfSwing, HammerThrow, HighJump, JavelinThrow, LongJump, PoleVault, Shotput\\
        \midrule
        \textit{\textbf{Validation:}} SoccerPenalty, TennisSwing\\
        \midrule
        \textit{\textbf{Testing:}} ThrowDiscus, VolleyballSpiking\\
        \bottomrule
    \end{tabularx}
    }
\end{table*}